\documentclass{article}

\usepackage{microtype}
\usepackage{graphicx}
\usepackage{subfigure}
\usepackage{booktabs} 
\usepackage{multirow}
\usepackage{amsmath, amssymb}
\usepackage{pst-all}
\usepackage{array, arydshln}
\usepackage{paralist}
\usepackage[format=hang]{caption}
\usepackage{algorithm}
\usepackage{algorithmic}
\newcommand{\R}{\mathbb{R}}

\newcommand{\doublecheckmark}{\checkmark$\!\!\!$\checkmark}

\DeclareMathOperator{\E}{E}
\DeclareMathOperator{\VAR}{VAR}

\usepackage{color}
\newcommand{\TODO}[1]{\textcolor{red}{#1}}
\renewcommand{\TODO}[1]{}  

\usepackage{hyperref}

\PassOptionsToPackage{numbers, sort&compress}{natbib}

\usepackage[final]{nips_2018}

\usepackage[utf8]{inputenc} 
\usepackage[T1]{fontenc}    
\usepackage{hyperref}       
\usepackage{url}            
\usepackage{booktabs}       
\usepackage{amsfonts}       
\usepackage{nicefrac}       
\usepackage{microtype}      

\title{Iteratively Training Look-Up Tables \\for Network Quantization}

\author{
  Fabien Cardinaux\thanks{Equal contribution: fabien.cardinaux@sony.com, stefan.uhlich@sony.com, kazuki.yoshiyama@sony.com}\\Sony Europe Ltd.\thanks{Sony European Technology Center, Stuttgart, Germany} \And Stefan Uhlich${}^*$\\Sony Europe Ltd.${}^\dag$ \And Kazuki Yoshiyama${}^*$ \\Sony Corporation\thanks{R\&D Center, Tokyo, Japan} \AND Javier Alonso García\\Sony Europe Ltd.${}^\dag$ \And Stephen Tiedemann\\Sony Europe Ltd.${}^\dag$ \And Thomas Kemp\\Sony Europe Ltd.${}^\dag$ \And Akira Nakamura\\Sony Corporation${}^\ddag$
}

\begin{document}

\maketitle

\setlength{\abovedisplayskip}{3pt}
\setlength{\belowdisplayskip}{3pt}

\begin{abstract}
Operating deep neural networks on devices with limited resources requires the reduction of their memory footprints and computational requirements. In this paper we introduce a training method, called \textit{look-up table quantization, LUT-Q}, which learns a dictionary and assigns each weight to one of the dictionary's values. We show that this method is very flexible and that many other techniques can be seen as special cases of LUT-Q. For example, we can constrain the dictionary trained with LUT-Q to generate networks with pruned weight matrices or restrict the dictionary to powers-of-two to avoid the need for multiplications. In order to obtain \textit{fully multiplier-less} networks, we also introduce a multiplier-less version of batch normalization. Extensive experiments on image recognition and object detection tasks show that LUT-Q consistently achieves better performance than other methods with the same quantization bitwidth.
\end{abstract}

\setdefaultleftmargin{1em}{2em}{}{}{}{}
\setlength{\itemsep}{0pt}

\section{Introduction and Proposed Training Method}
\label{sec:introduction}
\begin{figure*}
\vspace{-1.0cm}
\begin{minipage}[l]{0.60\linewidth}
    \hspace{-3.0cm}
		\resizebox{1.85\linewidth}{!}{
    \psset{arrowsize=2pt 3}
    \begin{pspicture}(0,-0.8)(20,4)
        \rput(3.30, 2.00){$\Bigg\{$}
        \rput(6.80, 2.00){$\Bigg\}$}
    
        \psframe(3.50,1.00)(3.70,3.00)
        \rput(3.60, 0.50){\small Dictionary}
        \rput(3.60, 0.15){\small $\mathbf{d} \in \R^K$}
        
        \psframe[fillstyle=hlines,hatchsep=0.025,hatchwidth=0.025,linewidth=0.01,linestyle=none](3.50, 1.90)(3.70, 2.10)
        \rput(3.60, 2.60){\tiny $\vdots$}
        \rput(3.60, 1.65){\tiny $\vdots$}

        \rput(3.90, 2.00){$,$}
        
        \psframe(4.60,1.00)(6.60,3.00)
        \rput(5.60, 0.50){\small Assignments}
        \rput(5.80, 0.15){\small $\mathbf{A} \in [1,\ldots,K]^{O\times I}$}

        \psframe[linewidth=0.01,linestyle=dashed](5.00, 1.00)(5.20, 3.00)
        \psframe[linewidth=0.01,linestyle=dashed](4.60, 2.30)(6.60, 2.50)
        
        \psframe[linewidth=0.01,linestyle=dashed](5.90, 1.00)(6.10, 3.00)
        \psframe[linewidth=0.01,linestyle=dashed](4.60, 1.30)(6.60, 1.50)
        
        \psline[arrows=->](7.00,2.00)(8.70,2.00)
        \rput(7.85, 2.25){\small\emph{Table lookup}}
        
        \psframe(8.90,1.00)(10.90,3.00)
        \rput(9.90,0.5){\small Weights (tied)}
        \rput(9.90,0.15){\small $\mathbf{Q} \in \R^{O\times I}$}
        \rput(9.90,-0.35){\small \emph{used in}}
        \rput(9.90,-0.7){\small \emph{forward/backward pass}}
        
        \psframe[linewidth=0.01,linestyle=dashed](9.30, 1.00)(9.50, 3.00)
        \psframe[linewidth=0.01,linestyle=dashed](8.90, 2.30)(10.90, 2.50)
        \psframe[fillstyle=hlines,hatchsep=0.025,hatchwidth=0.025,linewidth=0.01](9.30, 2.30)(9.50, 2.50)

        \psframe[linewidth=0.01,linestyle=dashed](10.20, 1.00)(10.40, 3.00)
        \psframe[linewidth=0.01,linestyle=dashed](8.90, 1.30)(10.90, 1.50)
        \psframe[fillstyle=hlines,hatchsep=0.025,hatchwidth=0.025,linewidth=0.01](10.20, 1.30)(10.40, 1.50)
        
        \rput(11.30,2.00){$\approx$}
        
        \psframe(11.70,1.00)(13.70,3.00) 
        \rput(12.70,0.5){\small Weights (full prec.)}
        \rput(12.70,0.15){\small $\mathbf{W} \in \R^{O\times I}$}
        \rput(12.70,-0.35){\small \emph{updated by}}
        \rput(12.70,-0.70){\small \emph{optimizer}}
        
        \psframe[linewidth=0.01,linestyle=dashed](12.10, 1.00)(12.30, 3.00)
        \psframe[linewidth=0.01,linestyle=dashed](11.70, 2.30)(13.70, 2.50)
        \psframe[fillstyle=hlines,hatchsep=0.025,hatchwidth=0.025,hatchcolor=gray,linewidth=0.01](12.10, 2.30)(12.30, 2.50)

        \psframe[linewidth=0.01,linestyle=dashed](13.00, 1.00)(13.20, 3.00)
        \psframe[linewidth=0.01,linestyle=dashed](11.70, 1.30)(13.70, 1.50)
        \psframe[fillstyle=hlines,hatchsep=0.025,hatchwidth=0.025,hatchcolor=gray,linewidth=0.01](13.00, 1.30)(13.20, 1.50)
        
        \psline[linestyle=dashed](12.70, 3.15)(12.70, 3.55)
        \psline[linestyle=dashed](12.70, 3.55)(3.60, 3.55)        
        \psline[arrows=->,linestyle=dashed](5.60, 3.55)(5.60, 3.05)
        \psline[arrows=->,linestyle=dashed](3.60, 3.55)(3.60, 3.05)
        \rput(8.9375,3.80){\small \emph{k-means ($M$ iterations)}}                       
    \end{pspicture}}
    \caption{Proposed look-up table\\quantization scheme.}
    \label{fig:weight_tying:sec:weight_tying_nets}
\end{minipage}
\hspace{-0.2cm}
\begin{minipage}[r]{0.39\linewidth}
    \centering
    \resizebox{1.0\linewidth}{!}{\includegraphics[trim=0 0 0 25,clip]{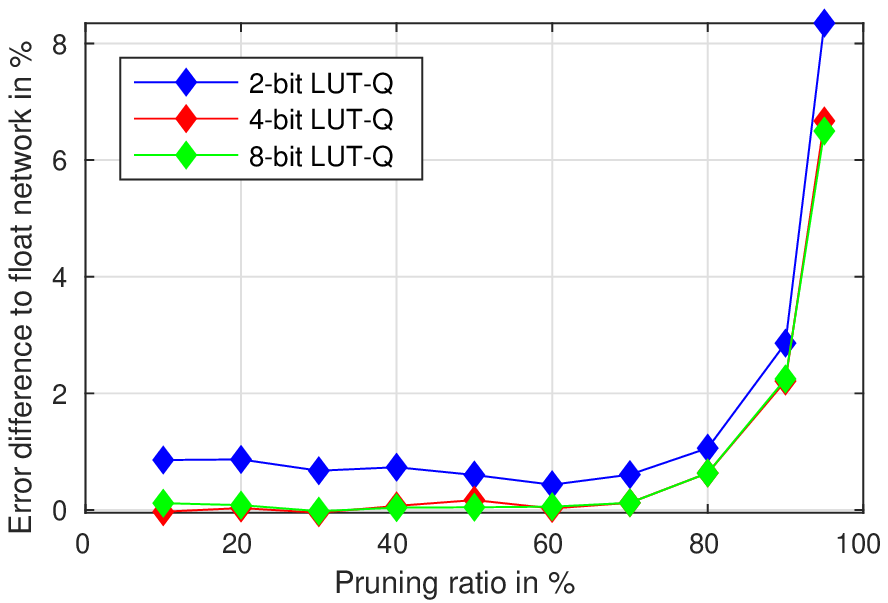}}
    \caption{CIFAR-10: Val. error for LUT-Q with pruning.}
    \label{fig:cifar10_pruning}
\end{minipage}
\end{figure*}

In this paper, we propose a training method for reducing the size and the number of operations of a \emph{deep neural network} (DNN) that we call \textit{look-up table quantization} (LUT-Q). As depicted in Fig.~\ref{fig:weight_tying:sec:weight_tying_nets}, LUT-Q trains a network that represents the weights $\mathbf{W}\in\R^{O\times I}$ of one layer by a dictionary $\mathbf{d}\in\R^K$ and assignments $\mathbf{A}\in[1, \ldots,K]^{O\times I}$ such that ${Q}_{oi} = d_{A_{oi}}$, i.e., elements of $\mathbf{Q}$ are restricted to the $K$ dictionary values in $\mathbf{d}$. To learn the assignment matrix $\mathbf{A}$ and dictionary $\mathbf{d}$, we iteratively update them after each minibatch. Our LUT-Q algorithm, run for each mini-batch, is summarized in Table~\ref{alg:WTN}.

LUT-Q has the advantage to be very flexible. By simple modifications of the dictionary $\mathbf{d}$ or the assignment matrix $\mathbf{A}$, it can implement many weight compression schemes from the literature. For example, we can constrain the assignment matrix and the dictionary in order to generate a network with pruned weight matrices. Alternatively, we can constrain the dictionary to contain only the values $\{-1,1\}$ and obtain a \textit{Binary Connect Network} \cite{courbariaux2015binaryconnect}, or to $\{-1,0,1\}$ resulting in a \textit{Ternary Weight Network} \cite{li2016ternary}. Furthermore, with LUT-Q we can also achieve \emph{Multiplier-less networks} by either choosing a dictionary $\mathbf{d}$ whose elements $d_k$ are of the form $d_k \in \{\pm2^{b_k}\}$ for all $k = 1,\ldots,K$ with $b_k \in \mathbb{Z}$, or by rounding the output of the $k$-means algorithm to powers-of-two. In this way we can learn networks whose weights are powers-of-two and can, hence, be implemented without multipliers.

The memory used for the parameters is dominated by the weights in affine/convolution layers. Using LUT-Q, instead of storing $\mathbf{W}$, the dictionary $\mathbf{d}$ and the assignment matrix $\mathbf{A}$ are stored. Hence, for an affine/convolution layer with $N$ parameters, we reduce the memory usage in bits from $N B_\text{float}$ to just $KB_\text{float} + N\left\lceil \log_2 K\right\rceil$, where $B_\text{float}$ is the number of bits used to store one weight. Furthermore, using LUT-Q we also achieve a reduction in the number of computations: for example, affine layers trained using LUT-Q need to compute just $K$ multiplications at inference time, instead of $I$ multiplications for a standard affine layer with $I$ input nodes.

\section{Experiments}
\label{sec:results}

For the description of our results we use the following naming convention: \emph{Quasi multiplier-less} networks avoid multiplications in all affine/convolution layers, but they are not completely multiplier-less since they contain multiplications in standard \emph{batch normalization} (BN) layers. For example, the networks described in~\cite{zhou2017incremental} are quasi multiplier-less. \emph{Fully multiplier-less} networks avoid all multiplications at all as they use our multiplier-less BN (see appendix~ \ref{sec:multiplierless_bn}). Finally, we call all other networks \emph{unconstrained}.

We conducted extensive experiments with LUT-Q and multiplier-less networks on the CIFAR-10 image classification task \cite{krizhevsky2009learning}, on the ImageNet ILSVRC-2012 task \cite{russakovsky2015imagenet} and on the Pascal VOC object detection task~\cite{everingham2009ThePV}. All experiments are carried out with the \textit{Sony Neural Network Library}\footnote{Neural Network Libraries by Sony: https://nnabla.org/}.

For CIFAR-10, we first use the full precision 32-bit \textit{ResNet-20} as reference ($7.4\%$ error rate). Quasi multiplier-less networks using LUT-Q achieve $7.6\%$ and $8.0\%$ error rate for 4-bit and 2-bit quantization respectively. Fully multiplier-less networks with LUT-Q achieve $8.1\%$ and $9.0\%$ error rates, respectively. LUT-Q can also be used to prune and quantize networks simultaneously. Fig.~\ref{fig:cifar10_pruning} shows the error rate increase between the baseline full precision \textit{ResNet-20} and the pruned and quantized network. Using LUT-Q we can prune the network up to $70\%$ and quantize it to 2-bit without significant loss in accuracy.

For Imagenet, we used \textit{ResNet-18}, \textit{ResNet-34} and \textit{ResNet-50} \cite{he2016deep} as reference networks. We report their validation error in Table~\ref{tab:IMAGENET-compared}. In Table~\ref{tab:IMAGENET-compared}, we compare LUT-Q against the published results using the INQ approach \cite{zhou2017incremental}, which also trains networks with power-of-two weights. We also compare with the baseline reported  \cite{mishra2018apprentice} which correspond the best results from the literature for each weight and quantization configuration. Note that we cannot directly compare the results of this \textit{appentrice} method \cite{mishra2018apprentice} itself because they do not quantize the first and last layer of the \textit{ResNets}. We observe that LUT-Q always achieves better performance than other methods with the same weight and activation bitwidth except for \textit{ResNet-18} with 2-bit weight and 8-bit activation quantization. Remarkably, \mbox{\textit{ResNet-50}} with 2-bit weights and 8-bit activations achieves $26.9\%$ error rate which is only $1.0\%$ worse than the baseline. The memory footprint for parameters and activations of this network is only $7.4$MB compared to $97.5$MB for the full precision network. Furthermore, the number of multiplications is reduced by two orders of magnitude and most of them can be replaced by bit-shifts.  

Finally, we evaluated LUT-Q on the Pascal VOC \cite{everingham2009ThePV} object detection task. 
We use our implementation of YOLOv2 \cite{redmon2017yolo9000} as baseline. This network has a memory footprint of 200MB and achieves a \textit{mean average precision (mAP)} of $72\%$ on Pascal VOC. We were able to reduce the total memory footprint by a factor of 20 while maintaining the mAP above $70\%$ by carrying out several modifications: replacing the feature extraction network with traditional residual networks \cite{he2016deep}, replacing the convolution layers by factorized convolutions\footnote{Each convolution is replaced by a sequence of pointwise, depthwise and pointwise convolutions (similarly to MobileNetV2 \cite{sandler2018mobilenetv2}}, and finally applying LUT-Q in order to quantize the weights of the network to 8-bit. Using LUT-Q with 4-bit quantization we are able to further reduce the total memory footprint down to just 1.72MB and still achieve a mAP of about $64\%$.

\begin{table*}
\hspace{-1.3cm}
\begin{minipage}[l]{0.5\linewidth}
   \caption{LUT-Q training algorithm}
   \label{alg:WTN}
\begin{algorithmic}
\begin{scriptsize}
   \STATE \textcolor{gray}{// Step 1:  Compute tied weights}
   \FOR{$l=1$ {\bfseries to} $L$}   
       \STATE $\mathbf{Q}^{(l)} = \mathbf{d}^{(l)}[\mathbf{A}^{(l)}]$
   \ENDFOR

    \vspace{0.15cm}

   \STATE \textcolor{gray}{// Step 2:  Compute current cost and gradients}
   \STATE $C=\text{Loss}\left(\mathbf{T},\text{Forward}\left(\mathbf{X},\mathbf{Q}^{(1)},\ldots,\mathbf{Q}^{(L)}\right)\right)$
   	\vspace{-1em}
		   \begin{align*}
		   \left\{\mathbf{G}^{(1)},\ldots,\mathbf{G}^{(L)}\right\} &= \left\{\frac{\partial C}{\partial\mathbf{Q}^{(1)}},\ldots,\frac{\partial C}{\partial\mathbf{Q}^{(L)}}\right\} \\
    &=\text{Backward}\left(\mathbf{X},\mathbf{T},\mathbf{Q}^{(1)},\ldots,\mathbf{Q}^{(L)}\right)\qquad\qquad\qquad
    		\end{align*}
	   
   \vspace{-0.17cm}
    
   \STATE \textcolor{gray}{// Step 3: Update full precision weights (here: SGD)}
   \FOR{$l=1$ {\bfseries to} $L$}
       \STATE  $\mathbf{W}^{(l)} = \mathbf{W}^{(l)} - \eta \mathbf{G}^{(l)}$
   \ENDFOR

   \vspace{0.15cm}
    
   \STATE \textcolor{gray}{// Step 4: Update weight tying by $M$ $k$-means iterations}
   \FOR{$l=1$ {\bfseries to} $L$}
       \FOR{$m=1$ {\bfseries to} $M$}
           \STATE $A^{(l)}_{ij} = \underset{k=1,\ldots,K^{(l)}}{\arg\min} \left\lvert W^{(l)}_{ij} - d^{(l)}_k \right\rvert$
           \FOR{$k=1$ {\bfseries to} $K^{(l)}$} 
           \STATE $d^{(l)}_k = \frac{1}{\sum\nolimits_{ij,\, A^{(l)}_{ij}=k} 1}\sum\nolimits_{ij,\, A^{(l)}_{ij}=k} W^{(l)}_{ij}$
           \ENDFOR
       \ENDFOR\vspace{-0.14cm}
   \ENDFOR
\end{scriptsize}
\end{algorithmic}
\end{minipage}
\hspace{-1.5cm}
\begin{minipage}[r]{0.50\linewidth}
\caption{ImageNet: LUT-Q compared to other quantization methods.\newline\doublecheckmark:~fully multiplier-less. \checkmark:~quasi multiplier-less. $\times$:~unconstrained.}
\label{tab:IMAGENET-compared}
\vskip -0.1in
\scriptsize
\setlength\extrarowheight{2pt}
\setlength\tabcolsep{0pt}
\begin{tabular}{rccccclcc}
\toprule
\multicolumn{3}{c}{\textbf{Quantization}}  && \multirow{2}{*}{\textbf{Source}} & \textbf{Multiplier-} & \multicolumn{3}{c}{\textbf{Validation error}}\\ 
\multicolumn{2}{c}{\textbf{Weights}} & \textbf{Activations} && 	                              &  \textbf{less}                       &  ResNet-18 & ResNet-34 & ResNet-50\\
\hline
32-bit &      & 32-bit  &&  our implementation             & $\times$  & 31.0\% & 28.1\% & 25.9\% \\
\hline
5-bit & pow-2 & 32-bit  &&  INQ \cite{zhou2017incremental} & \checkmark & 31.0\% & -      & 25.2\% \\
\hline
4-bit & pow-2 & 32-bit  &&  INQ \cite{zhou2017incremental} & \checkmark & 31.1\% & -      & -      \\
\hline
4-bit  &      & 8-bit   &&  \textit{apprentice} \cite{mishra2018apprentice}   & $\times$   & 33.6\% & 29.7\% & 28.5\% \\
4-bit & pow-2 & 8-bit   &&  \textbf{LUT-Q pow-2}       & \checkmark & 31.6\% & 28.1\% & 25.5\% \\
4-bit & pow-2 & 8-bit   &&  \textbf{LUT-Q pow-2}       & \doublecheckmark & 35.1\% & 30.7\% & 26.9\% \\
\hline
2-bit & pow-2 & 32-bit  &&  INQ \cite{zhou2017incremental} & \checkmark & 34.0\% & -      & -      \\
2-bit &       & 32-bit  &&  \textit{apprentice} \cite{mishra2018apprentice}    & $\times$   & 33.4\% & 28.3\% & 26.1\% \\
2-bit & pow-2 & 32-bit  &&  \textbf{LUT-Q pow-2}       & \checkmark & 31.8\% & -      & -      \\
\hline
2-bit   &     & 8-bit   &&  \textit{apprentice} \cite{mishra2018apprentice}    & $\times$   & 33.9\% & 30.8\% & 29.2\% \\
2-bit & pow-2 & 8-bit   && \textbf{LUT-Q pow-2}        & \checkmark & 35.8\% & 30.5\% & 26.9\% \\
2-bit & pow-2 & 8-bit   &&  \textbf{LUT-Q pow-2}       & \doublecheckmark & 43.2\% & 35.2\% & 29.8\%  \\
\bottomrule
\end{tabular}
\end{minipage}
\end{table*}

\section{Comparison to state-of-the-art}
\label{sec:related_work}

Different \emph{compression} methods were proposed in the past in order to reduce the memory footprint and the computational requirements of DNNs: pruning \cite{lecun1990handwritten, han2015learning}, quantization \cite{han2016deep, ullrich2017soft, chen2015compressing}, teacher-student network training \cite{romero2015fitnets, hinton2015distilling, mishra2018apprentice, polino2018model} are some examples. In general, we can classify the methods for quantization of the parameters of a neural network into three types:

\begin{compactitem}
    \item \emph{Soft weight sharing}: These methods train the full precision weights such that they form clusters and therefore can be more efficiently quantized~\cite{nowlan1992simplifying, chen2015compressing, ullrich2017soft, louizos2017bayesian, achterhold2018variational}.

	\item \emph{Fixed quantization}: These methods choose a dictionary of values beforehand to which the weights are quantized. Afterwards, they learn the assignments of each weight to the dictionary entries. Examples are \textit{Binary Neural Networks} \cite{courbariaux2015binaryconnect}, \textit{Ternary Weight Networks} \cite{li2016ternary} and also~\cite{mishra2017wrpn, mishra2018apprentice}.
		
    \item \emph{Trained quantization}: These methods learn a dictionary of values to which weights are quantized during training. However, the assignment of each weight to a dictionary entry is fixed \cite{han2016deep}.
\end{compactitem}

Our LUT-Q approach takes the best of the latter two methods: For each layer, we jointly update both dictionary and weight assignments during training. This approach to compression is similar to \emph{Deep Compression} \cite{han2016deep} in the way that we learn a dictionary and assign each weight in a layer to one of the dictionary's values using the $k$-means algorithm, but we update iteratively both assignments and dictionary at each mini-batch iteration.

\section{Conclusions and Future Perspectives}
\label{sec:conclusions_future_perspectives}

We have presented look-up table quantization, a novel approach for the reduction of size and computations of deep neural networks. After each minibatch update, the quantization values and assignments are updated by a clustering step. We show that the LUT-Q approach can be efficiently used for pruning weight matrices and training multiplier-less networks as well. We also introduce a new form of batch normalization that avoids the need for multiplications during inference. 

As argued in this paper, if weights are quantized to very low bitwidth, the activations may dominate the memory footprint of the network during inference. Therefore, we perform our experiments with activations quantized uniformly to 8-bit. We believe that a non-uniform activation quantization, where the quantization values are learned parameters, will help quantize activations to lower precision. This is one of the promising directions for continuing this work.

Recently, several papers have shown the benefits of training quantized networks using a \textit{distillation} strategy \cite{hinton2015distilling,mishra2018apprentice}. Distillation is compatible with our training approach and we are planning to investigate LUT-Q training together with distillation.

\bibliographystyle{icml2018}

\begin{thebibliography}{24}
\providecommand{\natexlab}[1]{#1}
\providecommand{\url}[1]{\texttt{#1}}
\expandafter\ifx\csname urlstyle\endcsname\relax
  \providecommand{\doi}[1]{doi: #1}\else
  \providecommand{\doi}{doi: \begingroup \urlstyle{rm}\Url}\fi

\bibitem[Achterhold et~al.(2018)Achterhold, Koehler, Schmeink, and
  Genewein]{achterhold2018variational}
Achterhold, J., Koehler, J.~M., Schmeink, A., and Genewein, T.
\newblock Variational network quantization.
\newblock \emph{International Conference on Learning Representations (ICLR)},
  2018.

\bibitem[Chen et~al.(2015)Chen, Wilson, Tyree, Weinberger, and
  Chen]{chen2015compressing}
Chen, W., Wilson, J., Tyree, S., Weinberger, K., and Chen, Y.
\newblock Compressing neural networks with the hashing trick.
\newblock In \emph{International Conference on Machine Learning (ICML)}, pp.\
  2285--2294, 2015.

\bibitem[Courbariaux \& Bengio(2016)Courbariaux and
  Bengio]{courbariaux2016binarynet}
Courbariaux, M. and Bengio, Y.
\newblock Binarynet: Training deep neural networks with weights and activations
  constrained to +1 or -1.
\newblock \emph{arXiv preprint arXiv:1602.02830}, 2016.

\bibitem[Courbariaux et~al.(2015)Courbariaux, Bengio, and
  David]{courbariaux2015binaryconnect}
Courbariaux, M., Bengio, Y., and David, J.-P.
\newblock Binaryconnect: Training deep neural networks with binary weights
  during propagations.
\newblock In \emph{Advances in Neural Information Processing Systems}, pp.\
  3123--3131, 2015.

\bibitem[Everingham et~al.(2009)Everingham, Van~Gool, Williams, Winn, and
  Zisserman]{everingham2009ThePV}
Everingham, M., Van~Gool, L., Williams, C. K.~I., Winn, J.~M., and Zisserman,
  A.
\newblock The pascal visual object classes (voc) challenge.
\newblock \emph{International Journal of Computer Vision}, 88:\penalty0
  303--338, 2009.

\bibitem[Han et~al.(2015)Han, Pool, Tran, and Dally]{han2015learning}
Han, S., Pool, J., Tran, J., and Dally, W.
\newblock Learning both weights and connections for efficient neural network.
\newblock In \emph{Advances in Neural Information Processing Systems}, pp.\
  1135--1143, 2015.

\bibitem[Han et~al.(2016)Han, Mao, and Dally]{han2016deep}
Han, S., Mao, H., and Dally, W.~J.
\newblock Deep compression: Compressing deep neural networks with pruning,
  trained quantization and {Huffman} coding.
\newblock In \emph{International Conference on Learning Representations
  (ICLR)}, 2016.

\bibitem[He et~al.(2016)He, Zhang, Ren, and Sun]{he2016deep}
He, K., Zhang, X., Ren, S., and Sun, J.
\newblock Deep residual learning for image recognition.
\newblock In \emph{Proceedings of the IEEE Conference on Computer Vision and
  Pattern Recognition}, pp.\  770--778, 2016.

\bibitem[Hinton et~al.(2015)Hinton, Vinyals, and Dean]{hinton2015distilling}
Hinton, G., Vinyals, O., and Dean, J.
\newblock Distilling the knowledge in a neural network.
\newblock In \emph{NIPS Deep Learning and Representation Learning Workshop},
  2015.

\bibitem[Ioffe \& Szegedy(2015)Ioffe and Szegedy]{ioffe2015batch}
Ioffe, S. and Szegedy, C.
\newblock Batch normalization: Accelerating deep network training by reducing
  internal covariate shift.
\newblock \emph{arXiv preprint arXiv:1502.03167}, 2015.

\bibitem[Krizhevsky \& Hinton(2009)Krizhevsky and
  Hinton]{krizhevsky2009learning}
Krizhevsky, A. and Hinton, G.
\newblock Learning multiple layers of features from tiny images.
\newblock 2009.

\bibitem[LeCun et~al.(1990)LeCun, Boser, Denker, Henderson, Howard, Hubbard,
  and Jackel]{lecun1990handwritten}
LeCun, Y., Boser, B.~E., Denker, J.~S., Henderson, D., Howard, R.~E., Hubbard,
  W.~E., and Jackel, L.~D.
\newblock Handwritten digit recognition with a back-propagation network.
\newblock In \emph{Advances in neural information processing systems}, pp.\
  396--404, 1990.

\bibitem[Li et~al.(2016)Li, Zhang, and Liu]{li2016ternary}
Li, F., Zhang, B., and Liu, B.
\newblock Ternary weight networks.
\newblock In \emph{NIPS Workshop on Efficient Methods for Deep Neural Networks
  (EMDNN)}, 2016.

\bibitem[Louizos et~al.(2017)Louizos, Ullrich, and
  Welling]{louizos2017bayesian}
Louizos, C., Ullrich, K., and Welling, M.
\newblock Bayesian compression for deep learning.
\newblock \emph{Conference on Neural Information Processing Systems (NIPS)},
  2017.

\bibitem[Mishra \& Marr(2018)Mishra and Marr]{mishra2018apprentice}
Mishra, A. and Marr, D.
\newblock Apprentice: Using knowledge distillation techniques to improve
  low-precision network accuracy.
\newblock \emph{International Conference on Learning Representations (ICLR)},
  2018.

\bibitem[Mishra et~al.(2017)Mishra, Nurvitadhi, Cook, and Marr]{mishra2017wrpn}
Mishra, A., Nurvitadhi, E., Cook, J.~J., and Marr, D.
\newblock Wrpn: Wide reduced-precision networks.
\newblock \emph{arXiv preprint arXiv:1709.01134}, 2017.

\bibitem[Nowlan \& Hinton(1992)Nowlan and Hinton]{nowlan1992simplifying}
Nowlan, S.~J. and Hinton, G.~E.
\newblock Simplifying neural networks by soft weight-sharing.
\newblock \emph{Neural computation}, 4\penalty0 (4):\penalty0 473--493, 1992.

\bibitem[Polino et~al.(2018)Polino, Pascanu, and Alistarh]{polino2018model}
Polino, A., Pascanu, R., and Alistarh, D.
\newblock Model compression via distillation and quantization.
\newblock \emph{International Conference on Learning Representations (ICLR)},
  2018.

\bibitem[Redmon \& Farhadi(2017)Redmon and Farhadi]{redmon2017yolo9000}
Redmon, J. and Farhadi, A.
\newblock Yolo9000: better, faster, stronger.
\newblock \emph{arXiv preprint}, 2017.

\bibitem[Romero et~al.(2015)Romero, Ballas, Kahou, Chassang, Gatta, and
  Bengio]{romero2015fitnets}
Romero, A., Ballas, N., Kahou, S.~E., Chassang, A., Gatta, C., and Bengio, Y.
\newblock Fitnets: Hints for thin deep nets.
\newblock In \emph{International Conference on Learning Representations
  (ICLR)}, 2015.

\bibitem[Russakovsky et~al.(2015)Russakovsky, Deng, Su, Krause, Satheesh, Ma,
  Huang, Karpathy, Khosla, Bernstein, et~al.]{russakovsky2015imagenet}
Russakovsky, O., Deng, J., Su, H., Krause, J., Satheesh, S., Ma, S., Huang, Z.,
  Karpathy, A., Khosla, A., Bernstein, M., et~al.
\newblock Imagenet large scale visual recognition challenge.
\newblock \emph{International Journal of Computer Vision}, 115\penalty0
  (3):\penalty0 211--252, 2015.

\bibitem[Sandler et~al.(2018)Sandler, Howard, Zhu, Zhmoginov, and
  Chen]{sandler2018mobilenetv2}
Sandler, M., Howard, A., Zhu, M., Zhmoginov, A., and Chen, L.-C.
\newblock Mobilenetv2: Inverted residuals and linear bottlenecks.
\newblock In \emph{The IEEE Conference on Computer Vision and Pattern
  Recognition (CVPR)}, June 2018.

\bibitem[Ullrich et~al.(2017)Ullrich, Meeds, and Welling]{ullrich2017soft}
Ullrich, K., Meeds, E., and Welling, M.
\newblock Soft weight-sharing for neural network compression.
\newblock In \emph{International Conference on Learning Representations
  (ICLR)}, 2017.

\bibitem[Zhou et~al.(2017)Zhou, Yao, Guo, Xu, and Chen]{zhou2017incremental}
Zhou, A., Yao, A., Guo, Y., Xu, L., and Chen, Y.
\newblock Incremental network quantization: Towards lossless {CNNs} with
  low-precision weights.
\newblock In \emph{International Conference on Learning Representations
  (ICLR)}, 2017.

\end{thebibliography}

\appendix
\newpage
\section{Multiplier-less Batch Normalization}
\label{sec:multiplierless_bn}
From~\cite{ioffe2015batch} we know that the traditional \emph{batch normalization} (BN) at inference time for the $o$th output is
\begin{equation}
	y_o \!=\! \gamma_o\frac{x_o - \E\left[x_o\right]}{\sqrt{\VAR\left[x_o\right]+\epsilon}} + \beta_o,
\label{eq:bn}
\end{equation}
where $\mathbf{x}$ and $\mathbf{y}$ are the input and output vectors to the BN layer, $\boldsymbol{\gamma}$ and $\boldsymbol{\beta}$ are parameters learned during training, $\E\left[\mathbf{x}\right]$ and $\VAR\left[\mathbf{x}\right]$ are the running mean and variance of the input samples, and $\epsilon$ is a small constant to avoid numerical problems. During inference, $\boldsymbol{\gamma}$, $\boldsymbol{\beta}$, $\E\left[\mathbf{x}\right]$ and $\VAR\left[\mathbf{x}\right]$ are constant and, therefore, the BN function \eqref{eq:bn} can be written as
\begin{equation}
	y_o = a_o \cdot x_o + b_o,
\label{eq:bn_simple}
\end{equation}
where we use the scale $a_o = \gamma_o / \sqrt{\VAR[x_o] + \epsilon}$ and offset $b_o = \beta_o - \gamma_o \E\left[x_o\right] / \sqrt{\VAR\left[x_o\right]+\epsilon}$. In order to obtain a multiplier-less BN, we require $\mathbf{a}$ to be a vector of powers-of-two during inference. This can be achieved by quantizing $\boldsymbol{\gamma}$ to $\boldsymbol{\hat\gamma}$. The quantized $\boldsymbol{\hat\gamma}$ is learned with the same idea as for WT: During the forward pass, we use traditional BN with the quantized $\boldsymbol{\hat\gamma} = \mathbf{\hat a} / \sqrt{\VAR[\mathbf{x}] + \epsilon}$ where $\mathbf{\hat a}$ is obtained from $\mathbf{a}$ by using the power-of-two quantization. Then, in the backward pass, we update the full precision $\boldsymbol{\gamma}$. Please note that the computations during training time are not multiplier-less but $\boldsymbol{\hat\gamma}$ is only learned such that we obtain a multiplier-less BN during inference time. This is different to \cite{courbariaux2016binarynet} which proposed a shift-based batch normalization using a different scheme that avoids all multiplications in the batch normalization operation by rounding multiplicands to powers-of-two in each forward pass. Their focus is on speeding up training by avoiding multiplications during training time, while our novel multiplier-less batch normalization approach avoids multiplications during inference.

\end{document}